\documentclass[10pt,letterpaper]{article}

\usepackage{cogsci}

\cogscifinalcopy 

\usepackage{pslatex}
\usepackage{apacite}
\usepackage{float} % Roger Levy added this and changed figure/table
\usepackage{graphicx}
\usepackage{mathtools}
\usepackage{booktabs}
\usepackage{tabularx}
\graphicspath{{pic/}}
\usepackage{enumitem}

\title{Deep Non-Monotonic Reasoning for Visual Abstract Reasoning Tasks}
 
\author{{\large \bf Yuan Yang (yuan.yang@vanderbilt.edu)} \\
{\large \bf Deepayan Sanyal (deepayan.sanyal@vanderbilt.edu)} \\
{\large \bf Joel Michelson (joel.p.michelson@vanderbilt.edu)} \\
{\large \bf James Ainooson (james.ainooson@vanderbilt.edu)} \\
{\large \bf Maithilee Kunda (mkunda@vanderbilt.edu)} \\
Vanderbilt University Department of Computer Science, 400 24th Ave S, Nashville, TN 37212 USA}

\begin{document}

\maketitle

\begin{abstract}
While achieving unmatched performance on many well-defined tasks, deep learning models have also been used to solve visual abstract reasoning tasks, which are relatively less well-defined, and have been widely used to measure human intelligence. However, current deep models struggle to match human abilities to solve such tasks with minimum data but maximum generalization. One limitation is that current deep learning models work in a monotonic way, i.e., treating different parts of the input in essentially fixed orderings, whereas people repeatedly observe and reason about the different parts of the visual stimuli until the reasoning process converges to a consistent conclusion, i.e., non-monotonic reasoning. This paper proposes a non-monotonic computational approach to solve visual abstract reasoning tasks. In particular, we implemented a deep learning model using this approach and tested it on the RAVEN dataset---a dataset inspired by the Raven's Progressive Matrices test. Results show that the proposed approach is more effective than existing monotonic deep learning models, under strict experimental settings that represent a difficult variant of the RAVEN dataset problem.

\end{abstract}

\section{Introduction}
Visual abstract reasoning (VAR) tasks have long played a prominent role in understanding and measuring human intellignece.  For instance, VAR task variants have been incorporated into many standard human intelligence tests, for example:  Wechsler, Stanford-Binet, Woodcock–Johnson, Kaufmann, etc. Psychometrics research has showed that VAR tasks often significantly load on core factors of intelligence, such as general intelligence and fluid intelligence within different theoretical models of human intelligence, such as Spearman's $g$, the three-stratum model of Cattell, Horn, \& Carroll, and the $g$-VPR model of Johnson \& Bouchard. 

A canonical example of VAR tasks is Raven's Progressive Matrices (RPM), which has been widely used in intelligence testing since the 1930s, and was shown to be the best single-format measure of general intelligence across a wide range of psychometric tests \shortcite{snow1984topography}. We thus use RPM-style matrix reasoning problems as being a representative and important variant of VAR tasks in the following discussion.

\begin{figure}[t]
    \centering
    \includegraphics[width=0.8\linewidth]{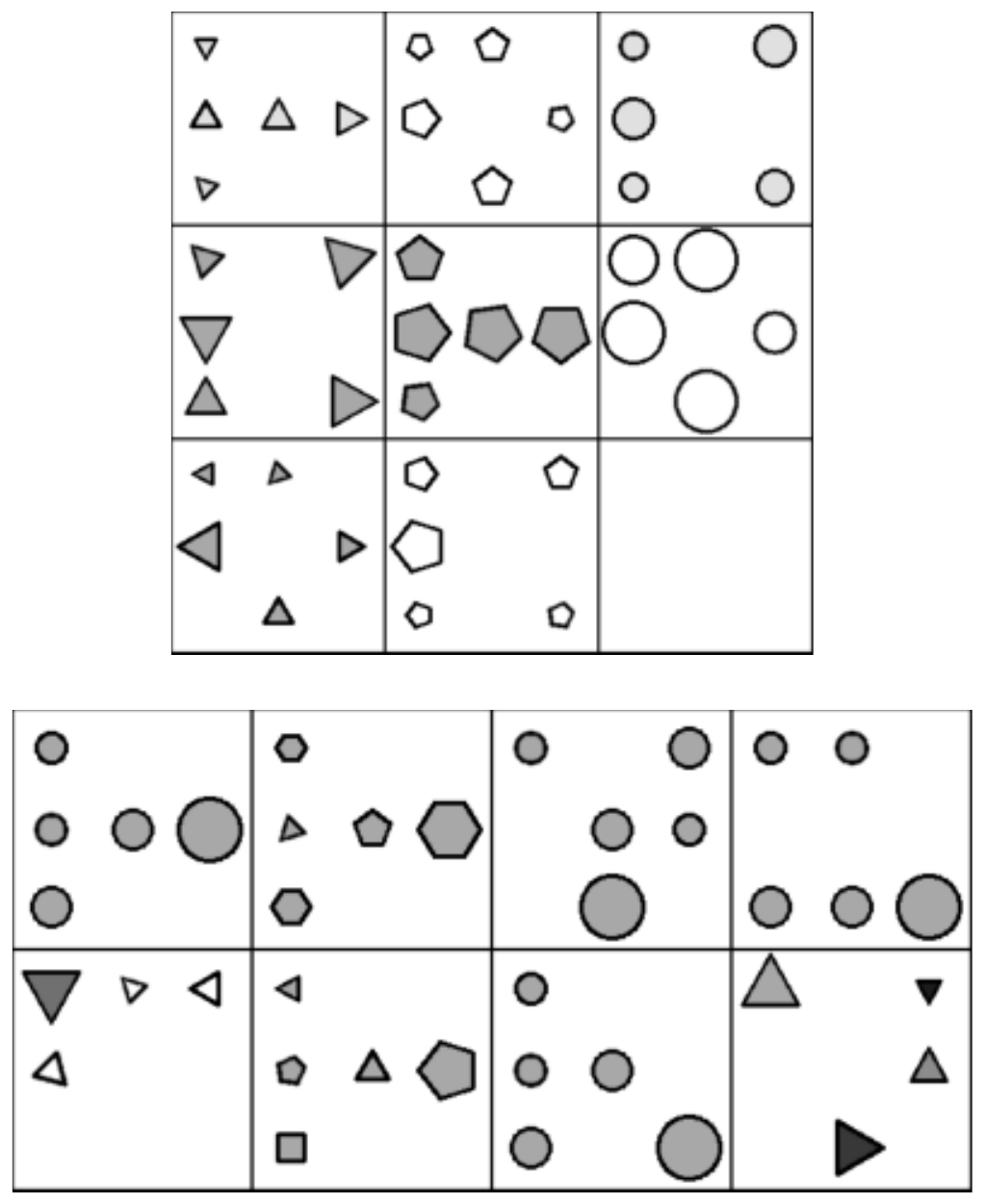}
    \caption{A sample item from the RAVEN dataset \shortcite{zhang2019raven}. The correct answer is the first image in the first row. It satisfies underlying four patterns in the matrix: (1) three spatial arrangements repeat in each row, but in different orders; (2) shapes in each column are the same; (3) the (relative) sizes of shapes in each row are constant; (4) two dark entries and one white entry are in each row.}
    \label{fig:rpm1}
\end{figure}

A matrix reasoning item consists of a matrix of visual figures with the last entry missing, along with several answer choices, with one answer choice that correctly completes the patterns in the matrix.  While the published RPM tests contain on the order of several dozen items manually designed by test author John Raven, recent efforts in AI has produced several large datasets of RPM-like problems to support machine learning research for solving VAR tasks.  Figure~\ref{fig:rpm1} shows an item from one such dataset called RAVEN that was inspired by the RPM \shortcite{zhang2019raven}.

% target: 0
% predict: 0
% meta_matrix:
% Distribute_Three Position
% Progression Type
% Constant Size
% Arithmetic Color
% meta_structure: [1 0 0 0 0 0 0 0 0 0 1 0 0 1 0 0 0 0 0 0 0]
% meta_target: [1 1 1 1 0 1 1 1 1]
% structure: [b'Scene' b'Singleton' b'Grid' b'Distribute_Nine' b'/' b'/' b'/' b'/']

Despite this extremely simple format, there is a lot we can tell about RPM. First, how could one determine that an RPM item is correctly solved? In other words, what are the criteria for an answer choice to be correct? Unfortunately, this is usually unclear until one sees the specific item. The designer of RPM items may use any patterns to create the items as long as the patterns can be visually expressed. The only rule is that, as intelligence tests, the underlying patterns should involve as less non-intelligence factors as possible, for example, languages and education experiences. This does not leave too many choices for item designers. In most cases, they can only use common geometric shapes, natural things (e.g., sun, moon, trees, and rivers), and body parts (hands and foot). Nonetheless, they created many novel and diversified designs for RPM tests and other similar intelligence tests. And more importantly, the correct answer is unambiguously favored by most human subjects when they find by themselves or are told the underlying patterns. Interesting! Isn't it? No universal criteria for being correct or wrong, but most people's understanding converges to the same answer. This requires much experience, especially when designing difficult items. For example, the item in Figure~\ref{fig:rpm1} looks very much in disarray at first glance. But, if one observes the underlying patterns of shape, color, and spatial arrangement (see the caption of Figure~\ref{fig:rpm1}), the first answer choice is obviously better than others. On the contrary, if a human subject has difficulty in any cognitive process related to extracting these underlying patterns, this item poses a great intellectual challenge.

Second, how are these RPM items solved differently by different people? There has been a long line of studies discussing various aspects of the human problem solving process, such as high-level strategies, e.g., constructive matching and response elimination \shortcite{bethell1984adaptive}, information processing components such as working memory capacity and selective attention \shortcite{carpenter1990one}, and perceptual complexity a\shortcite{primi2001complexity}. These works provide rich inspiration for building AI systems for solving RPM. Taking deep learning models as an example, these AI systems can be divided into single-choice and multiple-choice models, roughly corresponding to the constructive matching and response elimination strategies used by human subjects; some of these systems mimic the structure mapping process of analogy making to organize the information processing; and there also exists models with built-in memory and attention mechanisms, albeit not exactly the mechanisms of human cognition. However, these works, both in human and AI, are mostly focusing and stressing the importance of certain aspects or levels of the solving process. This has resulted in many ad hoc AI systems for solving RPM with less focus on flexible and integrative strategies. In this paper, we re-examine this question of more integrative solving processes for RPM-like tasks from scratch (i.e., with system inputs in the form of images at the pixel level).

Considering inputs at the pixel level poses an interesting challenge for AI systems in terms of inferential complexity:  If each entry in Figure~\ref{fig:rpm1} is a $160 \times 160$ gray scale image,  there exist 409,600 pixels in the image of the entire item. What pixels should an AI system process at each step of its computation? In what order, are these pixels processed? How are they processed? There are many possible answers to these questions. Eye tracking studies\shortcite{carpenter1990one} show that human subjects switch between matrix entries, rows, and columns during the entire solving process, but without a fixed order. Generally speaking, it is hard to tell when they are inspecting a matrix entry, when they are comparing multiple ones, and when they are drawing conclusions of the whole matrix. In contrast, every existing AI system for solving RPM assumes a certain order to process different parts of the visual input. These orders vary from simple ones, like sequentially processing every entry, to complex ones, like processing entries, pairs of entries, rows, columns, double rows, double columns, and combining the processings in a even more complex hierarchy.  In particular, these AI systems are designed with an assumption that the solving process---from pixel values to final predictions---is a monotonic one, where we use the term \textit{monotonic} to refer to a systematic ordering in how the reasoner processes elements of the problem input \shortcite{hoshen2017iq,barrett2018measuring,zhang2019learning,zhang2019raven,benny2021scale,zheng2019abstract,wang2020abstract,jahrens2020solving,shekhar2021neural,spratley2020closer,sinha2020memory,hahne2019attention,an2020hierarchical,rahaman2021dynamic,wu2021scattering,pekar2020generating,steenbrugge2018improving,van2019disentangled,kiat2020pairwise,hill2019learning}. 
While this assumption is effective for many deep learning tasks, it is limiting and not representative of human-like reasoning flexibility for solving VAR tasks. For example, the process could be non-monotonic by moving back and forth between perceptual and conceptual processings on different parts of raw visual input until it settles on a balanced point. Such a non-monotonic approach seems more reflective of human-like VAR reasoning, but such approaches have not received sustained attention in AI research on such tasks.

Again, take the RPM item in Figure~\ref{fig:rpm1} as an example. If we insert each answer choice into the matrix, we have 8 3$\times$3 matrices with each matrix entry a 160$\times$160 images. Since there are limited number of items in RPM-like datasets, we can assume that we have $N$ underlying patterns and each underlying pattern has $M$ instances (i.e., items) in this dataset. Then, this RPM-like dataset can be represented by a 7D array of shape $(N, M, a=8, r=3, c=3, h=160, w=160)$\footnote{$a$ for answer choices, $r$ for rows, $c$ for columns,  $h$ for height of matrix entry images, and $w$ for width of them.}. If we want our AI systems to learn the underlying patterns in the dataset, how does it process this huge array? If the $(h, w)$ and $(N)$ dimensions correspond to the lowest and highest cognitive levels, respectively, the the monotonic AI system would move the processing from the right to the left of $(N, M, c=8, r=3, c=3, h=160, w=160)$, without backtracking. The monotonic assumption here means that correctly deciphering a dimension is conditioned on correctly deciphering the ones on its right. Now, imagine a situation where all the entry images, and thus all the items, in the RPM-like dataset are the same. In this case, processing the first five dimensions first, rather than the last two, would be more efficient and easier for the AI system to detect the underlying pattern---everything is constant. This might be a very extreme example, but the general idea is clear. There should not be any fixed order to process input dimensions; at a certain point of processing for solving certain items, some dimensions are more informative than others. Therefore, the non-monotonic approach has its advantage over the monotonic approach as it sees all dimensions and allows to select the informative ones at each step.

Although the non-monotonic approach seems appealing, such freedom of processing poses another problem---we do not know what dimensions should be processed at each step and so many choices would make the model intractable. In this paper, we provide one solution to this problem:
\begin{enumerate}[nolistsep,noitemsep]
\item We describe a general non-monotonic computational approach for solving VAR tasks.
\item We implement a deep learning model to verify the feasibility of our non-monotonic approach.
\item We evaluate our model on the RAVEN dataset, using the strictest experimental settings that involve the least training data but the most underlying patterns, without any meta information of item generation or prior knowledge about the test, and using single-choice prediction, and we show that our model outperforms other deep learning models on this most challenging training and evaluation regime.
\end{enumerate}

\section{Related Work}

There is a long line of research of computational models solving VAR tasks, especially RPM-like tasks. A technical taxonomy classifies these works into four categories. We briefly describe them here.  More extensive reviews can be found in \shortcite{yang2022conceptual} and \shortcite{malkinski2022deep}.

\subsection{Imagery-Based Approach}
Visual mental imagery refers to mental images \shortcite{kosslyn2006case}. It plays a crucial role in human's ability to solve VAR tasks. The most important characteristic of mental imagery is that human can experience mental imagery in the absence of the concurrent sensory input. The imagery-based approach simulates human mental imagery by directly operating on the raw visual input and, through mental operations, such as rotation, addition, and subtraction, it can solve a substantial portion of original RPM tests \shortcite{yang2020not}.

\subsection{Logical Reasoning}
The computational models using logical reasoning works on symbolic representations of RPM-like items and reason in formal logic. For example, a entry image $A$ in a matrix could be described by a set of propositions: ``triangle($A$)=True, triangle-large($A$)=False, triangle-on-the-left($A$)=True, square($A$)=True, square-small($A$)=True, etc''. The symbolic representations in these models are either manually constructed or obtained through a preprocessing module. Representatives of this approach are ANALOGY \shortcite{evans1964program}, FAIRMAN ,and BETTERMAN \shortcite{carpenter1990one}.

\subsection{Neuro-Symbolic Approach}
The neuro-symbolic model usually consist of two modules---a neural perception frontend and a symbolic reasoning backend. The neural perception frontend (usually implemented as neural networks) extracts/approximates the distributions over the values of objects in each entry in a predefined formal representation system. The symbolic reasoning backend performs probability calculation according to a predefined set of rules. Neuro-symbolic approach and the next approach are data-driven approach, whereas the first two approaches are knowledge-based approaches. Examples of neuro-symbolic models for solving RPM-like tasks include ALANS2, PrAE, VAE-GPP, TRIVR, LoGe, and NVSA \shortcite{zhang2020learning, zhang2021abstract, shi2021raven, he2021two, yu2021abstract, hersche2022neuro}.  

\subsection{Learning Approach}
Unlike the previous approach, learning approach does not rely on predefined representation systems of geometric objects and abstract patterns. Instead, the representations are learned from raw perceptual data. This approach has become more popular because of large datasets of VAR tasks created in the last five years \shortcite{barrett2018measuring,zhang2019raven,hu2020stratified,benny2021scale}. When this paper is written, almost of all the popular deep learning architectures have been applied to solve VAR tasks, such as CNN, ResNet family, recurrent neural networks, and attention models. The proposed model in the following sections also falls into this category.

\section{Method}

\begin{figure*}[t]
    \centering
    \includegraphics[width=0.8\linewidth]{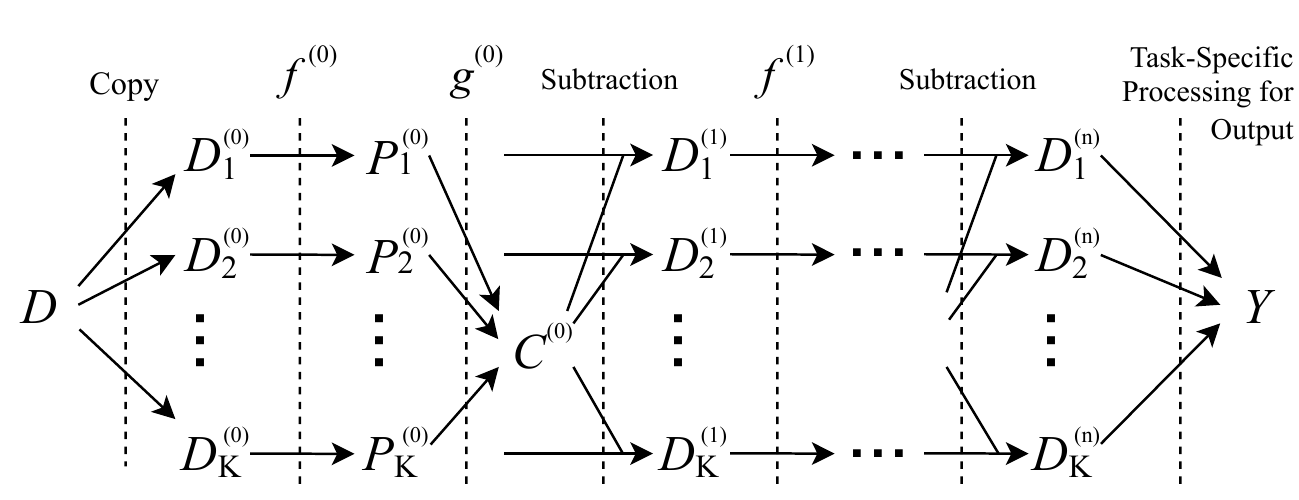}
    \caption{Non-monotonic processing of a multi-dimensional array $D$. This figure visualizes the computation of Equation~\eqref{eq:pcc0}, \eqref{eq:pcc1}, \eqref{eq:pcc2}, and \eqref{eq:pcc3}.}
    \label{fig:pc}
\end{figure*}

Let's continue the example of the RPM item in the introduction section, which is a 7D array of shape $(N, M, a=8, r=3, c=3, h=160, w=160)$. Each dimension represent a distinct type of information about this huge input array (although the information in different dimensions might be entangled). For example, the $(r, c)$ dimensions are used to realize the underlying patterns; the difference between answer choices, i.e., how well they complete the underlying patterns, is realized through the $a$ dimension; and different types of underlying patterns are organized in the $N$ dimension. All the first five dimensions are related to some information that is more abstract than the last two dimensions $(h,w)$. Thus, we refer to them as conceptual and perceptual dimensions, respectively. To further simplify the discussion, we consider the 4D sub-array of shape $(r=3, c=3, h=160, w=160)$, which represents a single matrix completed by an (correct or wrong) answer choice, with $(h=160, w=160)$ and $(r=3, c=3)$ as perceptual and conceptual dimensions, respectively. The task is then simplified to taking as input such a 4D array and providing as output indicating whether or not the answer choice is the correct one. So, it becomes a binary classification problem.   

As mentioned above, monotonic systems process perceptual and conceptual dimensions sequentially, whereas non-monotonic systems are allowed to process in any order. Generally, if we want to process conceptual dimensions for $a$ times and perceptual dimensions for $b$ times, there are ${a+b \choose a}$ ways to perform these processings sequentially. Since none of them is guaranteed to be more effective than others, we can try to perform all of them in parallel.

But how is this different from using multiple monotonic systems simultaneously in parallel? Well, let's take a close look at our example of the 4D array. We first insert a trivial dimension as an abstract feature dimension and it is thus of shape $(r=3, c=3, h=160, w=160, f=1)$. Consider two possible processing paths.
\begin{itemize}
    \item Path 1: $(r=3, c=3, h=160, w=160, f=1) \xRightarrow{f_{h,w}} (r=3, c=3, h=1, w=1, f=16) \xRightarrow{f_{r,c}} (r=1, c=1, h=1, w=1, f=32)$
    \item Path 2: $(r=3, c=3, h=160, w=160, f=1) \xRightarrow{f'_{r,c}} (r=1, c=1, h=160, w=160, f=16) \xRightarrow{f'_{h,w}} (r=1, c=1, h=1, w=1, f=32)$
\end{itemize}
$f_{h,w}$ and $f'_{h,w}$ are perceptual processings on the $(h,w)$ dimensions, with $(r,c)$ dimensions being transparent to them. Similarly, $G_{r, c}$ and $G'_{r, c}$ are conceptual processings on the $(r,c)$ dimensions, with $(h,w)$ dimensions being transparent to them. In this case, Path 1 represents the model's bottom-up ``understanding'' of underlying patterns of the matrix, whereas Path 2 represents the model's top-down ``understanding'' of underlying pattern of the matrix. Which one is correct? There is no universal answer to this question. The only condition one can tell is that only when the outputs of the two paths are consistent (or even identical) with each other can they be correct\footnote{This is very common in optimization and machine learning models. For example, many optimization problem is optimized when it meets its dual problem, i.e., they are optimized simultaneously.}, for that two different understandings of the same thing are both correct is impossible \footnote{This is true when the RPM item is well-designed for intelligence testing. The ambiguous case of VAR tasks is beyond the scope of this paper.}. Therefore, to make non-monotonic approach work and work better than an ensemble of multiple monotonic processing paths in parallel, we can try to force the consistency between the multiple paths, through a contrasting module. 

Now, we formalize the non-monotonic approach. Suppose that we have an input array $D$ of shape $(d_1, d_2, ..., d_l)$. Let $\Omega_1, \Omega_2, ..., \Omega_T \subset \{ 1, 2, ..., l \}$ be subsets of dimensions that we are going to process sequentially. Thus, there exist $K=T!$ possible processing paths and we create $K$ copies $\{ D^{(0)}_1, D^{(0)}_2, ..., D^{(0)}_K \}$ of $D$, with each copy for a path. A non-monotonic step, say Step $j$, transforms $\{ D^{(j)}_1, D^{(j)}_2, ..., D^{(j)}_K \}$ to $\{ D^{(j + 1)}_1, D^{(j+1)}_2, ..., D^{(j+1)}_K \}$ through Equation~\eqref{eq:pcc0}, \eqref{eq:pcc1}, \eqref{eq:pcc2}, and \eqref{eq:pcc3}. For all path $i \in  \{ 1, 2, ..., K \}$:
\begin{align}
    & f^{(j)}_i = f^{(j)}_{\Omega_{i_T}} \circ f^{(j)}_{\Omega_{i_{T - 1}}} \circ \cdots \circ f^{(j)}_{\Omega_{i_1}}  \label{eq:pcc0}\\
    & P^{(j)}_i = f^{(j)}_i \left( D^{(j)}_i \right) \label{eq:pcc1} \\
    & C^{(j)} = g^{(j)} \left( P^{(j)}_1, P^{(j)}_2, ..., P^{(j)}_K \right) \label{eq:pcc2} \\
    & D^{(j+1)}_i = P^{(j)}_i- C^{(j)}  \label{eq:pcc3}
\end{align}
where $f^{(j)} = (f^{(j)}_1, f^{(j)}_2, ..., f^{(j)}_K)$ and $g^{(j)}$ are processing paths and contrasting module of Step $j$. The processing paths $f^{(j)}$ and the contrasting module $g^{(j)}$ can be implemented using standard deep learning modules, such as convolutions, linear, and pooling, or other custom operations, as long as the input and output dimensions are compatible with neighboring modules. 

\begin{figure*}[t]
    \centering
    \includegraphics[width=0.8\linewidth]{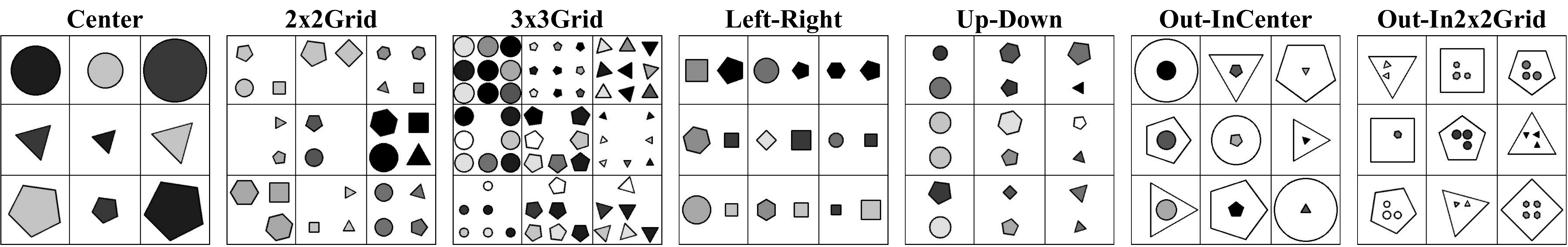}
    \caption{7 spatial configurations of the RAVEN dataset.}
    \label{fig:raven_configs}
\end{figure*}

Figure~\ref{fig:pc} depicts the non-monotonic approach in a more extended way. 
 In practice, we can stack multiple non-monotonic steps together, as we stack multiple convolutional layers in monotonic models.
At Step $0$, $f^{(0)}$ extracts information from certain dimensions of the input data array $D$ and optionally reduces the sizes of the dimensions; $g^{(0)}$ then contrasts the outputs from different paths. Since the parallel paths of  $f^{(0)}$ are processing the same dimensions in different orders, the outputs of different paths are expected to be consistent (or even identical) with each other, as we have discussed in the introduction section. Therefore, the contrasting module $g^{(0)}$ computes how the outputs from different paths are not consistent. Intuitively, the contrasting process will lead to two outcomes. First, if the outputs are consistent, the contrasting module will produce a certain type of output to give the following modules a signal that ``I have had a good understanding of the input'', and the following modules then respond to this signal by producing the desired final output, e.g., $Y=0$ in the Figure~\ref{fig:pc}. Second, if the outputs from different paths are not consistent, the contrasting module will filter out the components that causes the inconsistency, for which the following modules are more and more likely to be able to ``figure out'' why they are inconsistent, and, if enough non-monotonic steps are stacked, achieve the status of being consistent. To sum up, in either case, the model aims to achieve the consistency between different ways to decipher the raw input. 

\section{Experiment}

\subsection{Dataset}
To verify feasibility of the non-monotonic approach, we implemented a deep learning model to solve the RAVEN dataset \shortcite{zhang2019raven}---an RPM-like dataset that was procedurally generated according to 4 predefined underlying patterns (constant, progression, arithmetic, distribute of three) and 5 predefined geometric elements (number/position, type, size, color)\footnote{The values of geometric elements are limited in finite discrete sets. Thus, the RAVEN dataset is not as diversified as real-world intelligence tests.}. All the items in RAVEN share the same format in Figure~\ref{fig:rpm1}, i.e., a $3 \times 3$ matrix with 8 answer choices. The goemetric elements in RAVEN are organized in 7 predefined spatial configurations, as in Figure~\ref{fig:raven_configs}. Each configuration was used to generated 10,000 items. Thus, there are 70,000 items in total. We use the common splits of 60\%, 20\%, and 20\% for training, validation, and testing, respectively.

\subsection{Model Implementation}
Our model takes as input arrays of shape $(r=3, c=3, h=160, w=160)$, which represents a matrix completed by each of its answer choices, and provides as output the probability that the answer choice is correct. For each item, the answer choice with the highest probability is selected as the model's response to the item, which is used to compute the accuracy. 

\begin{table*}[t]
\centering
\caption{The implementation of our model. Note that the parameters``k'', ``s'', and ``p'' are kernel size, stride, and padding, respectively. The ``$\times$2'' imples that there are two data arrays of the shape, each being processed by a path, as shown in Figure~\ref{fig:pc}.}
\label{tab:model}
\begin{tabular}{c | c | l}
\toprule
Layer Name & Output Shape & Implementation \\
\midrule
     Input & (3$\times$3,1,160,160) & \\
     \hline
     Conv1 &  (3$\times$3,64,40,40) & Conv2d(1,64,k=7,s=2,p=3), BatchNorm2d(64), Relu, Maxpool2d(k=3,s=2,p=1) \\
     \hline
     Conv2 & (3$\times$3,64,20,20) & Conv2d(64,64,k=3,s=1,p=1), BatchNorm2d(64), Relu, Maxpool2d(k=3,s=2,p=1) \\
     \hline
     $f^{(0)}$ & (3,3,128,10,10)$\times$2 & \begin{tabular}{l}
          Path 1: ResidualBlock2d(64,64,k=3,s=1), ResidualBlock2d(64,128,k=3,s=2) \\
          Path 2: ResidualBlock2d(64,128,k=3,s=2), ResidualBlock2d(128,128,k=3,s=1) 
     \end{tabular}\\
     \hline
     $g^{(0)}$ & (3,3,128,10,10)$\times$2 & Avg of its two inputs, Conv3d(128,k=3,s=1,p=1),BatchNorm3d(128), Subtraction\\
     \hline
     $f^{(1)}$ & (3,3,256,5,5)$\times$2 & \begin{tabular}{l}
          Path 1: ResidualBlock2d(128,128,k=3,s=1), ResidualBlock2d(128,256,k=3,s=2) \\
          Path 2: ResidualBlock2d(128,256,k=3,s=2), ResidualBlock2d(256,256,k=3,s=1) 
     \end{tabular}\\
     \hline
     $g^{(1)}$ & (3,3,256,5,5)$\times$2 & Avg of its two inputs, Conv3d(256,k=3,s=1,p=1),BatchNorm3d(256), Subtraction\\
     \hline
     Pool & (256)$\times$2 & AvgPool4d(1) \\
     \hline
     MLP & (1) & Multi-Layer Perceptron of sizes [512, 256, 1] \\
\bottomrule
\end{tabular}
\end{table*}

Table~\ref{tab:model} describes the specific implementation of our model in the experiment. We use two convolutional layers before non-monotonic steps to reduce the sizes of $(h, w)$ dimensions, because they are too large and cost too much computation if we use only non-monotonic steps. In this model, we use two non-monotonic steps to process the $(r,c)$ and $(h,w)$ dimensions alternatively\footnote{The contrasting modules should have been 4D convolutions. But, it is too inconvenient to implement it in Pytorch and we used 3D convolutions instead.}. The model is implemented in Pytorch and trained with ADAM optimizer.

\begin{table*}[t]
\centering
\caption{Model Model Performance on RAVEN \shortcite{zhang2019raven, zhang2019learning}. CoPINet is evaluated on a variant of the RAVEN dataset since its result on the original RAVEN is not reliable\shortcite{hu2020stratified}.}
\label{tab:acc}
\begin{tabular}{c c c c c c c c c}
\toprule
Model & Avg. Acc. & Center & 2x2Grid & 3x3Grid & L-R & U-D & O-IC & O-IG \\
\midrule
     CNN & 36.97\% & 33.58\% & 30.30\% & 33.53\% & 39.43\% & 41.26\% & 43.20\% & 37.54\% \\
     ResNet & 53.43\%  & 52.82\% & 41.86\% & 44.29\% & 58.77\% & 60.16\% & 63.19\% & 53.12\% \\
     LSTM& 13.07\% &  13.19\% & 14.13\% & 13.69\% & 12.84\% & 12.35\% & 12.15\% & 12.99\%  \\
     WReN& 14.69\% & 13.09\% & 28.62\% & 28.27\% & 7.49\% & 6.34\% & 8.38\% & 10.56\%\\
     CoPINet & 46.1\% & 54.4\% & 36.8\% & 31.9\% & 52.2\% & 42.8\% & 51.9\% & 52.5\% \\
     Ours & \textbf{79.32\%} & \textbf{95.6\%} & \textbf{55.25\%} & \textbf{56.95\%} & \textbf{97.55\%} & \textbf{96.15\%} & \textbf{95.95\%} & \textbf{58.5\%} \\
\bottomrule
\end{tabular}
\end{table*}

\subsection{RAVEN Evaluation Settings}
Our experiment is by no mean comprehensive since there are several different variants of RAVEN\shortcite{yang2022automatic} and several different evaluation protocols (e.g., access to meta information of item generation, prior knowledge about the test, single-choice or multi-choice, and generation or decision). The combination of all these exponentially increases the cost of experiments, which we cannot afford in this single study. And the task difficulty also varies greatly in different combinations. Since our goal is to solve general VAR tasks, we chose the most ordinary one without using the distinctive features of RAVEN, i.e., the experimental setting of the least data but the most underlying patterns, without any meta information of item generation or prior knowledge about the test, and using single-choice prediction, in which the model needs to assess each answer choice separately, i.e., not allowed to compare them. Therefore, our experiment setting is the hardest one among all the settings for solving RAVEN, which has never been used before to the best of our knowledge. 

\subsection{Results}
Table~\ref{tab:acc} gives the accuracies of our model and other important baseline models on RAVEN. Our model achieves the best accuracies on both the entire dataset and each subset of spatial configuration. The results of WReN and CoPINet were obtained in similar settings to ours. The other three models---CNN, ResNet, and LSTM---are common deep learning models to provide a clue of how RAVEN compares to other deep learning tasks. Admittedly, there exist works, such as \shortcite{spratley2020closer, benny2021scale, hu2020stratified, wu2021scattering}, that can achieve higher performance than ours by using easier experimental settings, such as using meta information of item generation, such as using the underlying rules and geometric elements as additional supervising signal, and prior knowledge about RAVEN, such as decomposing a matrix into rows and processing each row individually (because the underlying patterns in RAVEN are represented only by rows). Because these works use significantly more input information and prior knowledge, we do not include them for direct comparison here.

Table~\ref{tab:acc} shows that our model does not work as well on the 2x2Grid, 3x3Grid, and Out-InGrid configurations, compared to other configurations. All three of these configurations contain grid structures, which allow more variation in number and position of geometric elements. These three configurations and other four configurations, respectively, represent two common types of perceptual organization in VAR tasks---classical view and normal view \shortcite{arendasy2005effect}---in which the correspondence between geometric elements are established in very different ways. Our model is good at one of them but not the other. However, it still achieves more than 50\% on its worst configurations, which is strong given the random baseline of 12.5\%. This raises an interesting question for further investigation---because the model is learning abstract patterns and already achieved 50\% accuracy, it must have learned some representations of the abstract concepts and be able to detect them in some cases; but what stops it from detecting them in other cases?

\section{Concluding Remarks}

Through RPM, we can see some common characteristics of visual abstract reasoning. First, there must be multiple parts in an VAR item because the abstract concepts are in nature relations between these multiple parts. In most cases, it requires additional dimensions to organize the multiple parts. The more complex the abstract concepts, the more additional dimensions needed. This nature of VAR task calls for effective multi-dimensional processing methods. The proposed non-monotonic approach is such a method. 

Second, the name ``visual abstract reasoning'' has been misleading for building AI systems as VAR tasks in intelligence tests are in no way pure reasoning problems. RPM tests and other VAR tests are for testing fluid intelligence, which is largely a superset of reasoning ability, or a precondition for the reasoning ability. This point can be seen through the administration procedures of RPM tests, in which very sparse instructions are given. The human subject only knows that she needs to point at one of the answer choices to complete the matrix, but have no idea of what ``complete'' means and what is the criterion of being correct or wrong. Therefore, VAR as intelligence tests is primarily about discovering and secondarily about reasoning. In contrast, the AI systems for solving VAR is mainly working on the reasoning part, with the discovering part ideally formulated by human designers. The proposed approach intends to capture the discovering part by pursuing the consistency between different processing paths. This approach is by no means the only way to do so, and it might not be a very efficient one. But the crux of building capable AI is definitely on the ``discovering'' part.

% The 50\% accuracies on dc, d9, and d4 is because of the optimization difficulty as exists in many other optimization processes, in which there exists two competing forces, such as GAN and VAE, and optima are achieved at a balance point. The competing forces in out model is the processing on different dimensions and the different orders to perform these processings.

\bibliographystyle{apacite}

\setlength{\bibleftmargin}{.125in}
\setlength{\bibindent}{-\bibleftmargin}
\newpage
\bibliography{main}

\end{document}